\documentclass{article} 
\usepackage{colm2024_conference}

\usepackage{microtype}
\usepackage{hyperref}
\usepackage{url}
\usepackage{booktabs}
\definecolor{darkblue}{rgb}{0, 0, 0.5}
\hypersetup{colorlinks=true, citecolor=darkblue, linkcolor=darkblue, urlcolor=darkblue}

\usepackage{graphicx}
\usepackage{soul}
\usepackage{tcolorbox}
\usepackage{multirow}
\usepackage{enumitem}
\usepackage{wrapfig}

\title{All in an Aggregated Image for In-Image Learning}

\author{\bf Lei Wang\textsuperscript{\rm1}$^\spadesuit$\hspace{3mm}
Wanyu Xu\textsuperscript{\rm2}$^\spadesuit$\hspace{3mm}
Zhiqiang Hu\textsuperscript{\rm3}$^\spadesuit$\hspace{3mm}
Yihuai Lan\textsuperscript{\rm4}$^\spadesuit$\hspace{3mm} \;
Shan Dong\textsuperscript{\rm1}$^\spadesuit$ \hspace{3mm}\\
\bf 
Hao Wang\textsuperscript{\rm4}\hspace{6mm}
Roy Ka-Wei Lee\textsuperscript{\rm3}\hspace{6mm}
Ee-Peng Lim\textsuperscript{\rm1}\hspace{4mm} \\
\normalsize{}\bf\textsuperscript{\rm1}Singapore Management University \\
\normalsize{}\bf\textsuperscript{\rm2}Southwest Jiaotong University\\
\normalsize{}\bf\textsuperscript{\rm3}Singapore University of Technology and Design\\
\normalsize{}\bf\textsuperscript{\rm4}The Hong Kong University of Science and Technology (Guangzhou)
}

\usepackage{amsmath,amsfonts,bm}

\def\eqref#1{equation~\ref{#1}}

\def\1{\bm{1}}

\DeclareMathAlphabet{\mathsfit}{\encodingdefault}{\sfdefault}{m}{sl}
\SetMathAlphabet{\mathsfit}{bold}{\encodingdefault}{\sfdefault}{bx}{n}

\colmfinalcopy 
\begin{document}

\maketitle

\begin{abstract}
This paper introduces a new in-context learning (ICL) mechanism called In-Image Learning (I$^2$L) that combines demonstration examples, visual cues, and chain-of-thought reasoning into an aggregated image to enhance the capabilities of Large Multimodal Models (e.g., GPT-4V) in multimodal reasoning tasks. 
Unlike previous approaches that rely on converting images to text or incorporating visual input into language models, I$^2$L consolidates all information into an aggregated image and leverages image processing, understanding, and reasoning abilities.
This has several advantages: it reduces inaccurate textual descriptions of complex images, provides flexibility in positioning demonstration examples, and avoids multiple input images and lengthy prompts. 
We also introduce I$^2$L-Hybrid, a method that combines the strengths of I$^2$L with other ICL methods. Specifically, it uses an automatic strategy to select the most suitable method (I$^2$L or another certain ICL method) for a specific task instance.
We conduct extensive experiments to assess the effectiveness of I$^2$L and I$^2$L-Hybrid on MathVista, which covers a variety of complex multimodal reasoning tasks. Additionally, we investigate the influence of image resolution, the number of demonstration examples in a single image, and the positions of these demonstrations in the aggregated image on the effectiveness of I$^2$L. Our code is publicly available at \url{https://github.com/AGI-Edgerunners/IIL}.
\end{abstract}

\section{Introduction}


Recently, there has been significant progress in large language models (LLMs)~\citep{brown2020language, chowdhery2022palm, openai-chatgpt-2022, touvron2023llama, openai-gpt4-2023}. The popularity of LLMs like ChatGPT~\citep{openai-chatgpt-2022} has inspired the development of various open-source LLMs~\citep{zhang2022opt, touvron2023llama, touvron2023llama2} and novel prompting methods~\citep{wei2022chain, zhou2022least, kojima2022large, zhang2022automatic, wang2023plan}.
Following ChatGPT, a powerful and versatile large multimodal model (LMM) with vision capabilities known as GPT-4V~\citep{openai2023gpt4vision} has been developed, which has shown strong abilities to understand both text and image inputs~\citep{yang2023dawn}.

Despite the success of GPT-4V, it still struggles with some multi-modal tasks~\citep{yang2023dawn}.
For example, when reading a complex plot, it may not be able to fully understand complex information presented in the image.
The underperformance of GPT-4V in these tasks could be due to not fully utilizing its capabilities~\citep{yang2023dawn, yang2023set, zhang2023lost}.  On the other hand,
\cite{yang2023dawn} find that GPT-4V can understand visual cues presented in images and propose Visual Referring Prompting~\citep{yang2023dawn} to manually edit input image pixels to incorporate visual cues such as arrows, boxes, and circles.
To further explore GPT-4V's capabilities in fine-grained visual grounding tasks, \cite{yang2023set} introduce Set-of-Mark prompting, which adds visual cues, like numeric or alphabetic marks, to specific image regions. Through set-of-mark prompting, GPT-4V can comprehend these regions more effectively.

\begin{figure}[t]
    \centering
    \includegraphics[width=1.0\linewidth]{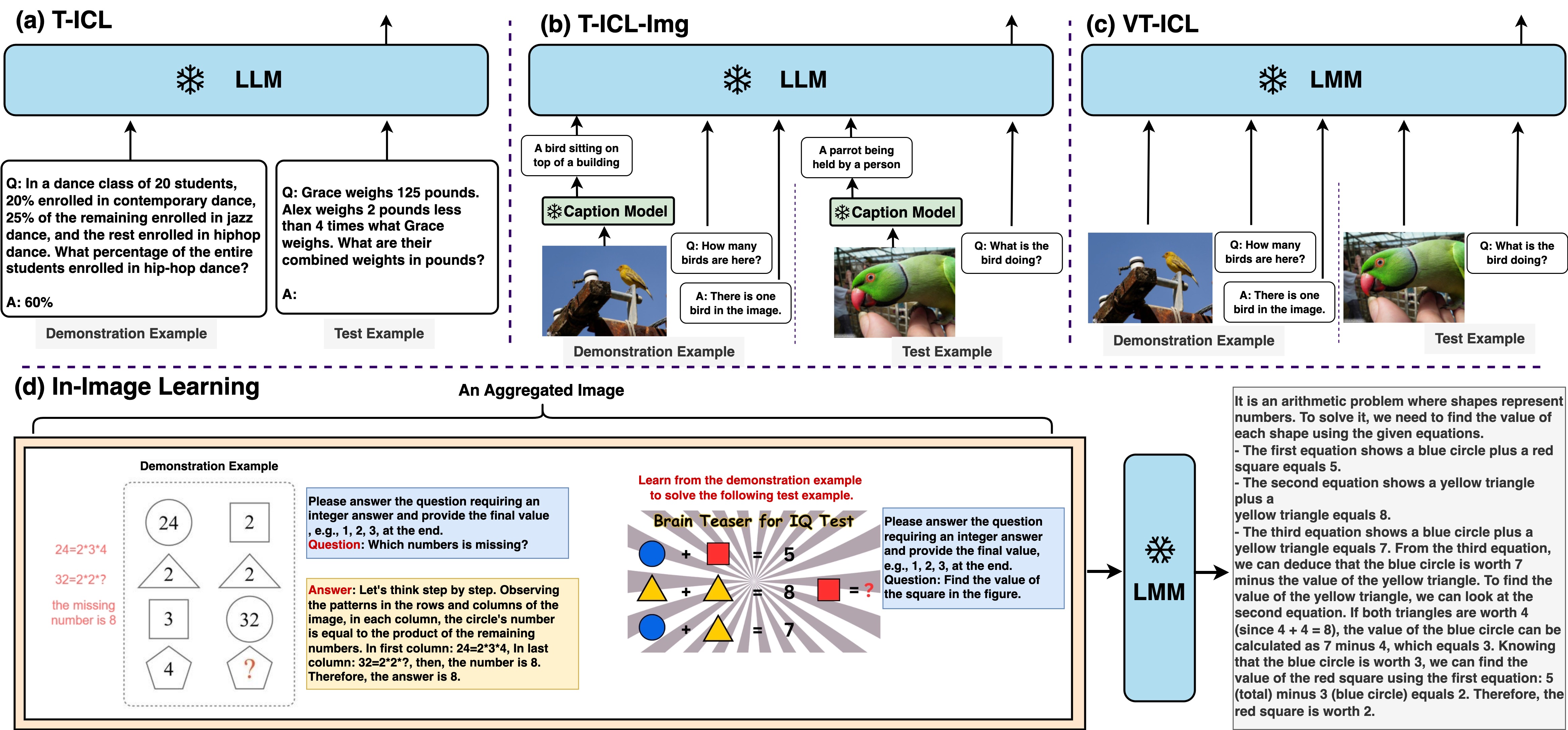}
    \caption{\textbf{(a)} Text-only in-context learning (T-ICL). \textbf{(b)} T-ICL with additional image-to-text models (T-ICL-Img). \textbf{(c)} Visual-text interleaved in-context learning (VT-ICL). \textbf{(d)} In-image learning (I$^2$L). For I$^2$L, we combine demonstrations (input image, visual cues, input text, output chain-of-thought reasoning, and output answer) and the test query (input image and input text), into an aggregated image. We then feed this aggregated image into LMMs to obtain the answer for the test query.}
    \label{fig:main}
    \vspace{-15pt}
\end{figure}

To further explore the potential of GPT-4V, in-context learning (i.e., learning from demonstration examples) is used to enhance its capabilities~\citep{yang2023dawn}.
In-context learning is widely used to help LLMs adapt to new natural language processing (NLP) tasks by learning from a few demonstration examples~\citep{brown2020language, liu2021makes, min2022rethinking, dong2022survey}. 
To transfer in-context learning from solving NLP tasks to multimodal tasks, a straightforward but widely used approach is to convert images into textual descriptions through additional image-to-text models~\citep{yang2022empirical, guo2022images}.
Flamingo~\citep{alayrac2022flamingo} eliminates the need for these additional models by directly encoding a series of interleaved visual-text demonstration examples into the developed visual language models one by one.

In this work, we introduce a new in-context learning mechanism called \textbf{I}n-\textbf{I}mage \textbf{L}earning (\textbf{I$^2$L}), which incorporates 
all useful information, which includes demonstrations (input image, visual cues, input text, output chain-of-thought reasoning, and output answer) and the test query (input image and input text), in an aggregated image to further unlock the reasoning ability of GPT-4V. 
In real-world scenarios with various tasks and data, it is infeasible to design and automatically add custom visual cues for each test data example.
Therefore, we manually create visual cues for demonstrations within each task, rather than adding them to test data examples for each task.
Consolidating valuable information into an aggregated image offers several advantages. Firstly, I$^2$L relies on image modeling, avoiding inaccurate textual description of complex images~\citep{hu2022promptcap}. 
Secondly, the input consists of only one image, eliminating the need for multiple images and lengthy prompt text. This reduces the overall input burden and costs of using LMMs~\citep{jiang2023longllmlingua, liu2023lost}.
Additionally, we observe that I$^2$L is good at handling complex images, while VT-ICL is better for images that can be easily described by text. To combine the strengths of these two methods for multimodal tasks, we propose I$^2$L-Hybrid using GPT-4V as an ICL method selector to determine the appropriate ICL method for each given multimodal task instance.

In our experiments, we conduct experiments on MathVista~\citep{lu2023mathvista} to test if GPT-4V can perform better on images, some of which are difficult to accurately describe with text alone, using I$^2$L.
Additionally, I$^2$L may be sensitive to image resolution, the number of demonstration examples in the aggregated image, and the position of demonstrations in the aggregated image. We thus experiment to further evaluate the impact of these factors on I$^2$L. In the Appendix, we also present experiments on three subsets of the VQA dataset~\citep{balanced_vqa_v2} and HallusionBench~\citep{liu2023hallusionbench} to verify the effectiveness of the proposed method.


\section{Related Work}
\label{sec:related_work}

\noindent\textbf{Prompting}
Recent advancements in the field of LLMs have received significant attention~\citep{brown2020language, chowdhery2022palm}, particularly following the success of models like ChatGPT~\citep{openai-chatgpt-2022}. This success has led to the development of various open-source LLMs~\citep{zhang2022opt, touvron2023llama, touvron2023llama2} and innovative prompting techniques~\citep{wei2022chain, zhou2022least, kojima2022large, zhang2022automatic, wang2023plan}.
Following ChatGPT, a powerful and versatile LMM called GPT-4~\citep{openai-gpt4-2023} with vision capabilities (GPT-4V~\citep{openai2023gpt4vision}) has been developed. GPT-4V can process and understand both textual and visual inputs, making it highly regarded for its ability to understand visual elements in images~\citep{yang2023dawn}. \cite{yang2023dawn} highlight GPT-4V's proficiency in recognizing and understanding visual signals, such as arrows, boxes, circles, and hand-drawn shapes, directly from images, and introduce visual referring prompting, which involves modifying image pixels to enhance visual cues. 
To unleash GPT-4V's ability to model fine-grained visual grounding, \citep{yang2023set} introduce a new prompting mechanism called \textit{Set-of-Mark (SoM) prompting}. This mechanism involves adding visual marks ((i.e., numeric or alphabetic labels))
to image regions so that GPT-4V can better understand and process these edited regions.
In this work, we expand on these approaches by integrating visual cues into demonstration examples rather than test queries. This allow us to leverage in-image learning and improve GPT-4V's performance on multimodal tasks, demonstrating its potential to understand and respond to complex inputs.

\noindent\textbf{In-Context Learning}

Recent advancements in large language models~\citep{brown2020language, chowdhery2022palm} have introduced a new capability, where models can adapt to a new NLP task by utilizing a few demonstration examples. This adaptation, known as in-context learning (ICL)~\citep{dong2022survey}, relies on learning from task-relevant demonstrations. The performance of ICL is greatly influenced by the wording of instructions~\citep{madaan2022text}, label design~\citep{yoo2022ground}, demonstration selection~\citep{liu2021makes, shi2022xricl} and ordering~\citep{lu2021fantastically}.
In the fields of multimodality, there have been early attempts at in-context learning. One example is Flamingo~\citep{alayrac2022flamingo}, which is trained by integrating visual inputs into LLMs. This enables the in-context learning of visual-linguistic tasks such as image captioning and OCR through language-based interfacing.
Otter~\citep{li2023otter}, a multi-modal model based on OpenFlamingo~\citep{awadalla2023openflamingo}, trained on a multi-modal in-context instruction tuning dataset and showcasing improved in-context learning ability.
To adapt LLMs from NLP tasks to more multimodal tasks, another common strategy is to leverage strong closed-source LLMs like ChatGPT, without pre-training and fine-tuning. This involves converting the corresponding images into textual descriptions, treating multimodal tasks as normal NLP tasks~\citep{yang2022empirical, guo2022images, hu2022promptcap, he2023icl}.
This work introduces a new in-context learning mechanism called In-Image Learning. It incorporates demonstrations and useful information into a single image, which is then fed into GPT-4V. Compare to previous in-context learning approaches for multimodal tasks, which convert images to textual descriptions and rely on the strong text processing ability of LLMs, in-image learning consolidates all information into one image and primarily leverages the image processing ability of LMMs.

\section{Visual In-Context Learning for Reasoning}
\label{sec:method}
Before we present the methods of visual in-context learning for reasoning, including our proposed methods, we first describe the existing methods like prompting with visual cues and text-only in-context learning (T-ICL). We then describe two state-of-the-art methods of visual in-context learning for reasoning, text-only in-context learning with additional image-to-text models (T-ICL-Img) and visual-text interleaved in-context learning (VT-ICL).  Finally, we introduce our proposed in-image learning (I$^2$L) and its variant in-image learning-hybrid (I$^2$L-Hybrid).

\subsection{Prompting with Visual Cues}
GPT-4V is good at understanding visual cues, such as symbols and numbers in the image. These cues are used in visual prompting methods, including Visual Referring Prompting~\citep{yang2023dawn} and Set-of-Mark prompting~\citep{yang2023set}. We augment the input image $x^{\text{img}}_q$ with visual cues by an operation $f_{\text{vc}}$ which edits the pixels of the corresponding image. The prediction result obtained by this prompting can be formulated as follows:
\begin{equation}
\underbrace{x^{\text{img}}_{\text{vc},q}, x^{\text{txt}}_q}_{\text {query }} \rightarrow \hspace{-0.3cm} \underbrace{\hat{y}_q^{\text{txt}}}_{\text {prediction }},
\end{equation}
where $x^{\text{img}}_{\text{vc},q} = f_{\text{vc}}\left(x^{\text{img}}_q\right)$ and $x^{\text{txt}}_q$ are the image with visual cues 
and text 
information of the test query $x_q$ respectively, and $\hat{y}_q^{\text{txt}}$ denotes the prediction generated by GPT-4V.

\subsection{Text-only In-Context Learning (T-ICL)}
Previous works have shown that T-ICL allows LLMs to solve tasks by learning from only a few demonstration examples~\citep{wei2022chain, dong2022survey}. In T-ICL, given $k$ text demonstration examples $x^{\text{txt}}_1, x^{\text{txt}}_2, \ldots, x^{\text{txt}}_k \in \mathcal{D}$ ($\mathcal{D}$ refers to the set of training ($x^{\text{txt}}_i$,$y^{\text{txt}}_i$) pairs) of a new task along with their corresponding ground truth labels $y^{\text{txt}}_1, y^{\text{txt}}_2, \ldots, y^{\text{txt}}_k$, the goal of ICL is to ask LLMs to generate $\hat{y}^{\text{txt}}_q$ as the predicted answer to a query data $x^{\text{txt}}_q$. T-ICL can be formulated as follows:
\begin{equation}
\underbrace{x^{\text{txt}}_1, y^{\text{txt}}_1, \ldots, x^{\text{txt}}_k, y^{\text{txt}}_k}_{\text {in-context examples (left-to-right) }}, \underbrace{x^{\text{txt}}_q}_{\text {query }} \rightarrow \hspace{-0.3cm}\underbrace{\hat{y}_q^{\text{txt}}}_{\text {prediction }}.
\end{equation}

\subsection{T-ICL-Img}
To adapt LLMs to a multimodal task, the T-ICL-Img strategy converts the task input image into textual description using some image-to-text models $f_{\text{i2t}}$~\citep{yang2022empirical, guo2022images}.
Formally, given $k$ input image-text data pairs $(x^{\text{txt}}_1, x^{\text{img}}_1), (x^{\text{txt}}_2, x^{\text{img}}_2), \ldots, (x^{\text{txt}}_{k}, x^{\text{img}}_{k}) \in \mathcal{D}$ with respective ground truth labels $y^{\text{txt}}_1, y^{\text{txt}}_2, \ldots, y^{\text{txt}}_{k}$, T-ICL-Img aims to output $\hat{y}_q^{\text{txt}}$ for the query $(x^{\text{txt}}_{q}, x^{\text{img}}_q)$ using LLMs based on the knowledge gained from the given $k$ data examples, formulated as:
\begin{equation}
\begin{split}
\underbrace{ x^{\text{cap}}_1, x^{\text{txt}}_1, y^{\text{txt}}_1, \ldots, x^{\text{cap}}_k, x^{\text{txt}}_k, y^{\text{txt}}_k}_{\text {in-context examples (left-to-right) }},  \underbrace{ x^{\text{cap}}_q, x^{\text{txt}}_q}_{\text {query }} \rightarrow \hspace{-0.3cm} \underbrace{\hat{y}_q^{\text{txt}}}_{\text {prediction }},
\end{split}
\end{equation}
where $x^{\text{cap}}_i = f_{\text{i2t}}\left(x^{\text{img}}_i\right)$ and $f_{\text{i2t}}$ represents the image-to-text model.

\subsection{VT-ICL} 
While the results of T-ICL-Img are promising, there is a potential risk of losing information when converting visual inputs into textual descriptions~\citep{yang2022empirical, hu2022promptcap}.
To avoid the need for additional image-to-text models, such as generic caption models, interleaved image-text pairs can be prepared for in-context learning, and visual inputs can be directly incorporated into the large vision-language model.
Formally, given $k$ input image-text data pairs $(x^{\text{txt}}_1, x^{\text{img}}_1), (x^{\text{txt}}_2, x^{\text{img}}_2), \ldots, (x^{\text{txt}}_k, x^{\text{img}}_k) \in \mathcal{D}$ with their ground truth labels $y^{\text{txt}}_1, y^{\text{txt}}_2, \ldots, y^{\text{txt}}_k$, VT-ICL aims to output $\hat{y}^{\text{txt}}_q$ for the query data $(x^{\text{txt}}_q, x^{\text{img}}_q)$ using LLMs based on the knowledge gained from the given $k$ demonstration examples, formulated as:
\begin{equation}
\underbrace{x^{\text{img}}_1, x^{\text{txt}}_1, y^{\text{txt}}_1, \ldots, x^{\text{img}}_k, x^{\text{txt}}_k, y^{\text{txt}}_k}_{\text {in-context examples (left-to-right) }}, \underbrace{x^{\text{img}}_q, x^{\text{txt}}_q}_{\text {query }} \rightarrow \hspace{-0.3cm} \underbrace{\hat{y}_q^{\text{txt}}}_{\text {prediction }}
\end{equation}

\subsection{Proposed I$^2$L}
This section presents the proposed In-Image Learning (I$^2$L), which combines visual-text demonstration examples, visual cues, instructions, and chain-of-thought reasoning into an aggregated image to enhance the capabilities of GPT-4V.
Consolidating valuable information into a single image offers two main benefits. Firstly, it effectively conveys complex images that cannot be accurately described by text alone. 
Thirdly, using only one image as input reduces the need for lengthy input, thereby reducing the input burden and avoiding exceeding the input limits of LMMs.

For each query, I$^2$L includes $k$ demonstrations to perform in-image learning.  Formally, to predict the output $\hat{y}^{\text{txt}}_q$ for a query $(x^{\text{img}}_q, x^{\text{txt}}_q)$, I$^2$L first convert each input image-text demonstration pair $(x^{\text{img}}_i,x^{\text{txt}}_i)$ and its ground truth labels $y^{\text{txt}}_i$ into an image containing $x^{\text{img}}_i$ augmented with visual cues and combined with the corresponding input text $x^{\text{txt}}_i$ and ground truth answer $y^{\text{txt}}_i$.  We denote this resultant image by $z_{\text{vc},i}^{\text{img}} = f_{vc}(x^{\text{img}}_{i},x^{\text{txt}}_{i},y^{\text{txt}}_{i})$.  Next, we combine $z_{\text{vc},i}^{\text{img}}$'s of all $k$ demonstrations into one single image $z_{all}^{\text{img}}$ in some position permutation order $\pi$ using an operation $f_{comb,\pi}$. Next, I$^2$L combines $z_{all}^{\text{img}}$ together with the query $(x^{\text{img}}_q, x^{\text{txt}}_q)$ into an image $z_{\text{comb}}$ by another operation $f_{comb}$.  Finally, we feed $z_{\text{comb}}$ to the LMM and obtain the prediction result.

Formally, we capture the above steps as follows:
\begin{equation}
\underbrace{z_{\text{comb}}}_{\substack{\text{demonstrations and}\\\text{query in an image}}} \rightarrow \hspace{-0.3cm} \underbrace{\hat{y}_q^{\text{txt}}}_{\text{prediction}}
\end{equation}
where 
$z_{\text{comb}}=f_{comb}(z_{all}^{\text{img}}, x^{\text{img}}_q, x^{\text{txt}}_q)$, and
$z_{all}^{\text{img}}=f_{comb,\pi}(z_{\text{vc},1}^{\text{img}}, \cdots, z_{\text{vc},k}^{\text{img}})$. 

Note that we do not include any visual cues or other annotations for the test query image $x^{\text{img}}_{q}$. 
In contrast, visual prompting methods like visual referring prompting and set-of-mark prompting do add visual cues to images of test data examples. Visual referring prompting manually adds these cues to the test query, but this method is challenging to scale up. On the other hand, set-of-mark prompting uses off-the-shelf interactive segmentation models to divide an image into regions and overlay these regions with marks such as alphanumerics. However, this approach relies on the object segmentation ability of the segmentation models and may not always be correct. 
Considering the above issues with visual prompting methods, I$^2$L introduces the addition of visual cues to demonstrations instead of test data examples.

\begin{figure}[t]
    \centering
    \includegraphics[width=0.7\linewidth]{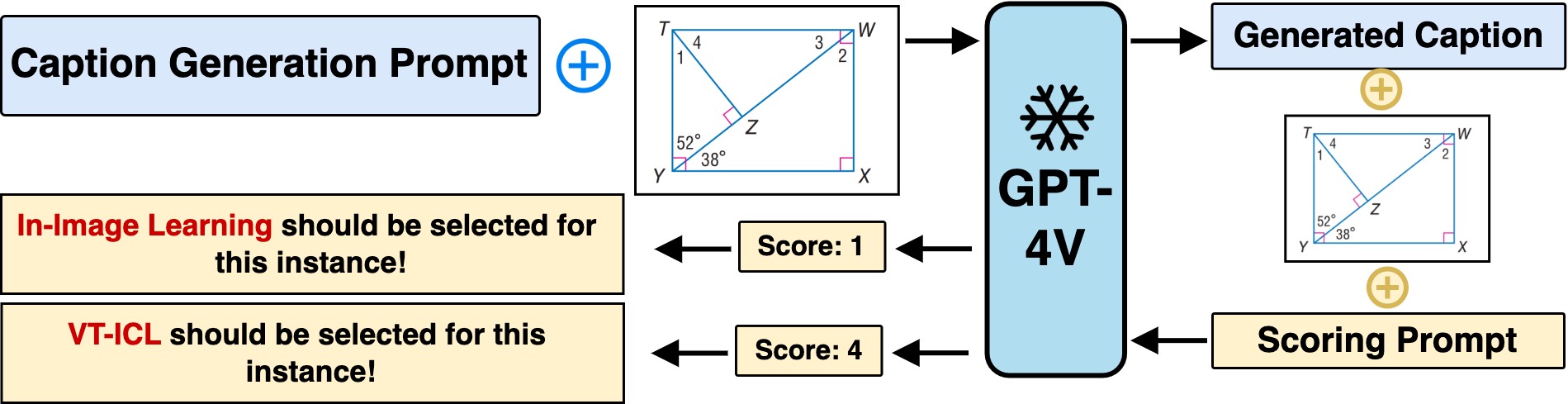}
    \vspace{-5pt}
    \caption{Overview of \textit{I$^2$L-Hybrid}.}
    \vspace{-15pt}
    \label{fig:selection}
\end{figure}

\subsection{I$^2$L-Hybrid}

The proposed I$^2$L excels at handling complex images that cannot be accurately described using text alone, while VT-ICL can better leverage text information to enhance performance for images that can be easily described by text. 
To combine the strengths of I$^2$L and VT-ICL methods for multimodal tasks, we draw inspiration from the \textit{GPT-4V-as-a-Generalist-Evaluator}~\cite{zhang2023gpt} and propose using GPT-4V as a selector called \textit{GPT-4V-Selector} to determine the appropriate method for a specific multimodal data example in a given task.

Specifically, we first prompt GPT-4V to generate a description for a given data example's image. The prompt for generating a description is shown as follows:
\begin{tcolorbox}
\texttt{\footnotesize
Describe the given image. 
\footnotesize$<$IMAGE$>$
}
\end{tcolorbox}
$<$IMAGE$>$ is a placeholder representing the image input for GPT-4V.

Then, we ask GPT-4V to rate this data example from 0 to 1 based on the comparison between the generated description and the image. The prompt for rating is shown below:

\begin{tcolorbox}
\texttt{\footnotesize
The following is a description of the current image. Please rate it from 1 to 4 based on the comparison between this description and the image, with a higher score for more accurate descriptions and a lower score for more vague descriptions. Your answer MUST be in the format: `The rating is [YOUR SCORE]'. Please be CLEAR and DO NOT provide any additional outputs or information beyond the requested output.\\
\footnotesize$<$DESCRIPTION$>$$<$IMAGE$>$
}
\end{tcolorbox}
$<$DESCRIPTION$>$ is a placeholder representing the description generated by the previous step. It will be used as input for GPT-4V, along with the instruction and the image. A lower rating score generated by GPT-4V indicates that the image is difficult to describe from GPT-4V's perspective, making it more suitable for I$^2$L. Conversely, a higher score suggests that the image is easy to describe, making VT-ICL more suitable for this data example. We have a threshold to determine whether the score is suitable for I$^2$L or VT-ICL.
Figure~\ref{fig:selection} provides an overview of the selection process. 

Formally, the ICL selector of I$^2$L-Hybrid is formulated as:
\begin{equation}
  \parbox{3em}{I$^2$L-Hybrid} =
    \begin{cases}
      \text{I$^2$L} & \text{if $f_{score}(x^{img}_q, \hat{x}^{cap}_q) \leq \theta$}\\
      \text{VT-ICL} & \text{otherwise}
    \end{cases}       
\end{equation}
where $\hat{x}^{cap}_q = LMM(x^{img}_q)$.

\section{Experiments}


In this section, we evaluate our proposed I$^2$L and I$^2$L-Hybrid on the MathVista~\citep{lu2023mathvista} benchmark to validate the effectiveness of I$^2$L and I$^2$L-Hybrid in enhancing LMM's comprehension and reasoning of images to answer questions.
The following sections will detail the methods used to create in-image demonstrations and the results of our experiments on MathVista. 

\subsection{Experiment Setup}
\textbf{Datasets.} 
We use the MathVista testmini \citep{lu2023mathvista} for multimodal reasoning tasks.
MathVista testmini comprises 1,000 instances from 28 multimodal datasets, including 
IQTest, FunctionQA, and PaperQA. These tasks requires the models to understand complex visual details and to perform complex reasoning, both are non-trivial for advanced LMMs. 

\noindent\textbf{Implementations.}
For all the experiments, we use GPT4-V as the backbone model, which is one of the most powerful VLMs with public APIs. We report the results of the \texttt{gpt4-vision-preview} engine with temperature equaling to 0. The \texttt{gpt4-vision-preview} engine is also used as the caption model for the T-ICL-Img method. In the context of few-shot experiments, the demonstration examples are drawn from a distinct set of samples for each benchmark. For instance, the demonstration examples are sampled from MathVista's entire test set, excluding the testmini set specifically utilized within this paper.


\textbf{Baselines.}
We include the baselines employed in MathVista \citep{lu2023mathvista}, where the baselines are under three setups: Text-Only LLMs in zero-shot and two-shot settings with CoT and Program-of-Thought (PoT), (b) Augmented-LLMs where the LLMs are provided with additional visual information including the generated image captions, and (c) LMMs that include open-source models.

For the T-ICL-Img baseline, we utilized GPT-4V to generate textual descriptions for both the demonstration images and test images. As for VT-ICL, the approach involves incorporating both images and textual demonstrations as input listed in an interleaved format.

\subsection{Demonstration Construction}

\subsubsection{Aggregated Image Construction for I$^2$L}
In this section, we explain how to construct the aggregated image for I$^2$L. This aggregated image includes demonstrations, each of which consist of an input image, visual cues, input text, output chain-of-thought reasoning, and an output answer, as well as a test query consisting of an input image and input text. We detail our design for the visual cues included in the demonstrations below.

\noindent\textbf{Visual Cues.}
In real-world scenarios with various tasks and data, it is infeasible to design and automatically add custom visual cues for each test data example.
Therefore, we manually create visual cues for demonstrations within each task, rather than adding them to test data examples for each task.
We hope these demonstrations with visual cues can guide GPT-4V in solving test data examples for each task.
The main idea behind creating visual cues is to emphasize critical elements and annotate necessary information to answer the given question.
Specifically, we highlight critical elements by adding bounding boxes in conspicuous colors. If necessary, we provide concise descriptions for these elements near the bounding boxes. 
If the text includes information absent from the objects or their relationships, we add this information close to the objects.
For example, in demonstrations involving bar charts, we include numerical values on the bars. In geometric problems, we assign numerical values to critic angles and line segments.
Additionally, we include the guiding sentence ``Learn from the demonstration examples to solve the following test example'' in the aggregated image. This explicitly instructs GPT-4V to learn from demonstrations and visual cues to solve test data examples.

Figure \ref{fig:I2L_mathvista} in the Appendix illustrates an example of aggregated image construction on the MathVista dataset. We incorporate visual cues, such as boxes, to highlight critic components within the example. Two bounding boxes are employed to identify the lowest value points and middle value points, and the calculation of the sum of these points is included within the image. Moreover, concise description for critical objects within the chart are provided. It is imperative during the integration of visual cues and demonstrations to ensure that no essential objects in the original image are obscured, thus mitigating the risk of information loss. Additionally, a Chain-of-Thought rationale is included to enrich GPT-4V's reasoning. Next, we combine the aforementioned image with a test data example to generate the final aggregated image. By learning from this demonstration example in the aggregated image, GPT-4V can more accurately solve the test data example included in the aggregated image.


\begin{table*}[t!]
\centering
 \small
 \renewcommand\tabcolsep{2.5pt} 
 \resizebox{0.98\linewidth}{!}{
    \begin{tabular}{l|c|ccccc|ccccccc}
    \toprule
    \bf{Model}  & \bf{ALL} & \bf{FQA} & \bf{GPS} & \bf{MWP} & \bf{TQA} & \bf{VQA} & \bf{ALG} & \bf{ARI} & \bf{GEO} & \bf{LOG} & \bf{NUM} & \bf{SCI} & \bf{STA}  \\ 
    \midrule
    \multicolumn{13}{l}{ \textit{ Heuristics  baselines}} \\
    \midrule
    Random chance  & 17.9 & 18.2 & 21.6 & 3.8 & 19.6 & 26.3 & 21.7 & 14.7 & 20.1 & 13.5 & 8.3 & 17.2 & 16.3 \\
    Frequent guess  & 26.3 & 22.7 & 34.1 & 20.4 & 31.0 & 24.6 & 33.1 & 18.7 & 31.4 & 24.3 & 19.4 & 32.0 & 20.9 \\
    \midrule
    \multicolumn{13}{l}{ \textit{LLMs} (Input: $x^{\text{txt}}$)} \\
    \midrule
    Zero-shot ChatGPT~\citep{openai-chatgpt-2022} & 23.5 & 21.9 & 26.9 & 9.1 & 38.6 & 23.5 & 27.7 & 15.9 & 25.7 & 21.6 & 9.9 & 41.5 & 20.5 \\
    Zero-shot GPT-4~\citep{openai-gpt4-2023}  & 26.1 & 22.3 & 37.0 & 7.0 & 39.2 & 27.4 & 33.6 & 17.4 & 35.6 & 16.2 & 9.2 & 45.8 & 19.5 \\
    Zero-shot Claude-2~\citep{claude2} & 26.4 & 21.9 & 34.1 & 13.4 & 36.1 & 29.1 & 32.8 & 20.4 & 33.3 & 13.5 & 12.1 & 36.4 & 20.5 \\
    \midrule
    2-shot CoT Claude-2~\citep{wei2022chain}   & 24.4 & 18.6 & 29.8 & 9.7 & 33.5 & {34.1} & 29.2 & 19.0 & 28.0 & 5.4 & 13.9 & 36.9 & 18.9 \\
    2-shot CoT ChatGPT~\citep{wei2022chain}   & 26.8 & 20.1 & 36.5 & 8.6 & 44.9 & 28.5 & 35.6 & 17.0 & 33.5 & 21.6 & 14.6 & 45.9 & 17.9 \\
    2-shot CoT GPT-4~\citep{wei2022chain}  & {29.2} & 20.1 & {44.7} & 8.6 & {46.2} & 31.3 & {41.6} & 19.3 & {41.0} & 18.9 & 13.9 & 47.5 & 18.9 \\
    \midrule
    2-shot PoT ChatGPT~\citep{chen2022program}  & 25.1 & 19.0 & 30.8 & {16.1} & 38.0 & 25.7 & 29.9 & 19.8 & 29.3 & {24.3} & {19.4} & 38.5 & 16.9 \\
    2-shot PoT GPT-4~\citep{chen2022program}  & 26.0 & 20.1 & 33.2 & 8.1 & 44.9 & 28.5 & 32.7 & 16.7 & 31.0 & {24.3} & 13.2 & {48.4} & 18.3 \\
    \midrule
    \multicolumn{13}{l}{ \textit{Augmented-LLMs} (Input: $x^{\text{txt}}$, $x^{\text{cap}}$)} \\
    \midrule
    2-shot CoT Claude-2~\citep{wei2022chain}  & 33.2 & 26.0 & 31.7 & 35.5 & 48.1 & 30.2 & 32.4 & 32.3 & 33.0 & 16.2 & 17.4 & 54.9 & 36.2 \\
    2-shot CoT ChatGPT~\citep{wei2022chain} & 33.2 & 27.5 & 29.3 & {36.0} & 49.4 & 29.1 & 31.0 & {32.9} & 31.0 & 16.2 & 17.4 & 50.8 & 37.2 \\
    2-shot CoT GPT-4~\citep{wei2022chain} &  33.2 & 27.9 & 31.7 & 31.2 & {51.9} & 28.5 & 33.5 & 30.9 & 32.2 & 13.5 & 12.5 & {58.2} & {37.9} \\
    \midrule
    2-shot PoT ChatGPT~\citep{chen2022program}  & 26.8 & 24.5 & 26.4 & 23.7 & 33.5 & 27.9 & 27.8 & 26.1 & 28.0 & {18.9} & 13.2 & 33.6 & 29.9 \\
    2-shot PoT GPT-4~\citep{chen2022program}  & {33.9} & {30.1} & {39.4} & 30.6 & 39.9 & {31.3} & {37.4} & 31.7 & {41.0} & {18.9} & {20.1} & 44.3 & {37.9} \\
    \midrule
    \multicolumn{13}{l}{\textit{LMMs} (Input: $x^{\text{txt}}$, $x^{\text{img}}$)} \\
    \midrule
    IDEFICS-9B-Instruct~\citep{laurencon2023obelics}  & 19.8 & 21.6 & 21.1 & 6.5 & 25.9 & 24.0 & 22.1 & 15.0 & 19.8 & 18.9 & 9.9 & 24.6 & 18.1 \\
    mPLUG-Owl-LLaMA-7B~\citep{ye2023mplug}  & 22.2 & 22.7 & 23.6 & 10.2 & 27.2 & 27.9 & 23.6 & 19.2 & 23.9 & 13.5 & 12.7 & 26.3 & 21.4 \\
    miniGPT-4-LLaMA-2-7B~\citep{zhu2023minigpt}  & 23.1 & 18.6 & 26.0 & 13.4 & 30.4 & 30.2 & 28.1 & 21.0 & 24.7 & 16.2 & 16.7 & 25.4 & 17.9 \\
    LLaMA-Adapter-V2-7B~\citep{zhang2023llama_adp}  & 23.9 & 21.2 & 25.5 & 11.3 & 32.3 & 31.8 & 26.3 & 20.4 & 24.3 & {24.3} & 13.9 & 29.5 & 18.3 \\
    LLaVAR~\citep{zhang2023llavar}  & 25.2 & 21.9 & 25.0 & 16.7 & 34.8 & 30.7 & 24.2 & 22.1 & 23.0 & 13.5 & 15.3 & 42.6 & 21.9 \\
    InstructBLIP-Vicuna-7B~\citep{instructblip}  & 25.3 & 23.1 & 20.7 & 18.3 & 32.3 & 35.2 & 21.8 & 27.1 & 20.7 & 18.9 & {20.4} & 33.0 & 23.1 \\
    LLaVA-LLaMA-2-13B~\citep{liu2023llava}  & 26.1 & 26.8 & 29.3 & 16.1 & 32.3 & 26.3 & 27.3 & 20.1 & 28.8 & {24.3} & 18.3 & 37.3 & 25.1 \\
    Multimodal Bard~\citep{google2023bard} & 34.8 & 26.0 & 47.1 & 29.6 & 48.7 & 26.8 & 46.5 & 28.6 & 47.8 & 13.5 & 14.9 & 47.5 & 33.0 \\
    {GPT-4V (Playground)~\citep{openai2023gpt4vision}} & {49.9} & {43.1} & \bf{50.5} & \underline{57.5} & {65.2} & {38.0} & \underline{53.0} & {49.0} & \bf{51.0} & 21.6 & 20.1 & {63.1} & {55.8} \\
    \midrule
    \multicolumn{13}{l}{ \textit{Our Implementation} (GPT-4V)} \\
    \midrule
    SoM Prompting (0-shot) &31.5 &{23.1} &{32.6} &{26.3} &{51.9} &{30.1} &{35.2} &{26.0} &{31.7}  &{16.2} &{20.1} &{52.0} &{26.0} \\
    T-ICL-Img (1-shot) & {49.1} & {45.3} & {49.0} & {56.4} & {61.7} & {36.8} & {49.8} & {47.3} & {50.2} & {21.6} & \underline{27.7} & {63.8} & {57.5} \\
    VT-ICL w/o visual cues (1-shot)  & \underline{51.6} & {49.8} & {42.3} & {60.7} & {65.6} & \underline{44.1} & {46.9} & \bf{53.2} & {43.9} & {24.3} & {29.1} & {67.2} & \underline{61.2} \\
    VT-ICL w/ visual cues  (1-shot) & \underline{51.6} & \underline{50.1} & {48.5} & {56.9} & \underline{65.8} & {39.4} & {51.6} & {48.3} & {48.9} & \underline{27.0} & {25.6} & \underline{67.5} & \underline{61.2} \\
    I$^2$L (1-shot) & {51.5} & {49.6} & {40.8} & \bf{58.0} & \bf{67.7} & \bf{45.8} & {46.9} & \underline{50.9} & {43.9} & \bf{29.7} & \bf{32.6} & {65.2} & {59.0} \\
    \hline
    I$^2$L-Hybrid (I$^2$L, VT-ICL w/ visual cues)& \bf{52.8} & \bf{51.6} & \underline{50.4} & \bf{58.0} & \underline{65.8} & {41.0} & \bf{53.3} & {49.5} & \underline{50.6} &\bf{29.7} & {25.7} & \bf{68.0} & \bf{62.5} \\
    \midrule
    \multicolumn{13}{l}{ \textit{Human}} \\
    \midrule
    Human performance & 60.3 & 59.7 & 48.4 & 73.0 & 63.2 & 55.9 & 50.9 & 59.2 & 51.4 & 40.7 & 53.8 & 64.9 & 63.9 \\
    \bottomrule
    \end{tabular}
    }
    \vspace{-5pt}
    \caption{Accuracy on the \textit{testmini} subset of MathVista (1000 test data examples). 
    ALL: overall accuracy. Task types: FQA: figure question answering, GPS: geometry problem solving, MWP: math word problem, TQA: textbook question answering, VQA: visual question answering. Mathematical reasoning types: ALG: algebraic reasoning, ARI: arithmetic reasoning, GEO: geometry reasoning, LOG: logical reasoning, NUM: numeric commonsense, SCI: scientific reasoning, STA: statistical reasoning. The highest and second highest scores across all models are bolded and underlined, respectively.}
\vspace{-5pt}
\label{tab:mathvista}
\end{table*}





\subsection{Main Results}
Table \ref{tab:mathvista} shows the performance results of various models on the MathVista \textit{testmini} dataset. All our implementations use a one-shot evaluation, designed to compare the effectiveness of different in-context learning paradigms. Notably, I$^2$L achieves an average accuracy of 51.5\% on MathVista, almost matching VT-ICL's 51.6\%. On the other hand, T-ICL-Img, which solely relies on textual input from GPT-4V, attains a lower average accuracy of 49.1\%. This highlights the potential information loss during the caption generation process for T-ICL-Img, even when using the GPT-4V model.

Performance variation across different MathVista subsets is apparent. T-ICL-Img shows superior performance in the GPS and GEO subsets. VT-ICL excels in the FQA, ALG, SCI, MWP, ARI and STA subsets. In contrast. I$^2$L outperforms other methods in the TQA, VQA, LOG, and NUM subsets. 
The images within subsets, TQA, VQA,LOG, and NUM, present challenges in description, yet the model can effectively learn from visual cues within demonstration examples to better grasp the patterns involved in solving such problems. Moreover, the integration of visual cues and Chain-of-Thought rationales can enhance the reasoning capabilities of GPT4-V, enabling it to address problems within these subsets.
The proposed I$^2$L-Hybrid method, combining the benefits of I$^2$L and VT-ICL, shows promising results.
By using GPT-4V as a selector, I$^2$L-Hybrid achieves the highest average accuracy of 52.8\% on MathVista.


\begin{figure*}[t]
    \centering
    \includegraphics[width=1\linewidth]{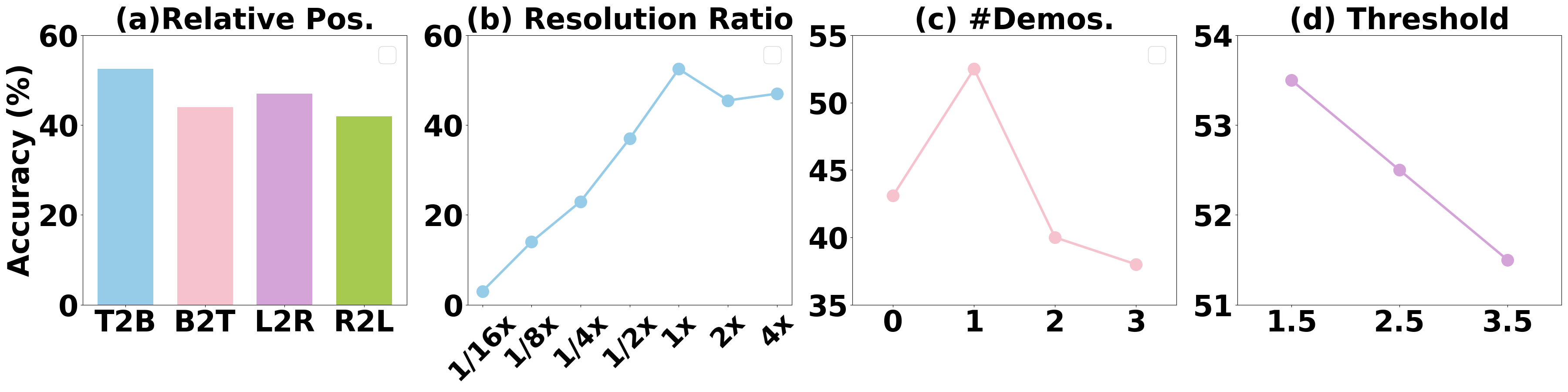}
    \vspace{-15pt}
    \caption{In-depth analysis for I$^2$L: (a) Impact of relative position of demonstrations and test examples in an aggregated image. T2B represents ``Top to Bottom'', meaning arranging the examples from top to bottom in sequence. B2T represents from bottom to top, L2R and R2L represent from left to right and from right to left. (b) Impact of resolution ratio. (c) Impact of the number of demonstrations. (d) Impact of thresholds for I$^2$L-Hybrid.}
    \label{fig:vqa_analysis}
    \vspace{-5pt}
\end{figure*}



\begin{table}[]
    \centering
    \resizebox{0.32\linewidth}{!}{
    \begin{tabular}{lcc}
    \toprule  
    Method& Acc.& \\
    \midrule  
    I$^2$L w/o chain-of-thought& {49.0} & \\
    I$^2$L w/o visual cues& {49.0} & \\
    \midrule
    I$^2$L& \textbf {52.5}& \\
    \bottomrule 
    \end{tabular}
    }
    \vspace{-10pt}
    \caption{An ablation study of the impact of different components in the proposed I$^2$L on a subset of 200 data examples from MathVista.}
    \vspace{-15pt}
    \label{tab:ablation}
\end{table}

\subsection{Analysis}

In this section, we conduct in-depth analyses on the proposed I$^2$L using a subset of 200 data examples from MathVista. Specifically, we conduct experiments to examine the impact of various factors, including the components of I$^2$L, resolution ratio, the number of demonstrations, and the positioning of the demonstration.

\noindent\textbf{Ablation Study.} Table~\ref{tab:ablation} demonstrates that adding chain-of-thought as the target in the image demonstration and incorporating visual cues in the image both contribute to the final performance. This suggests that I$^2$L  has the potential to enhance GPT-4V's ability by incorporating additional useful and relevant information into the aggregated image.

\noindent\textbf{Impact of Relative Position of Demonstrations and test examples.} Figure \ref{fig:vqa_analysis}(a) explores the performance variation based on the positioning of demonstration and test examples. The results show that I$^2$L is sensitive to the relative positions of demonstrations and test examples, and performs most effectively when the demonstrations and test examples are arranged from top to bottom.

\noindent\textbf{Impact of Resolution Ratio.} Figure \ref{fig:vqa_analysis}(b) illustrates the performance variation in response to changes in resolution ratio of an aggregated image. Notably, the model's performance tends to decline as the resolution ratio increases or decreases.

\noindent\textbf{Impact of Number of Demonstrations.} Figure \ref{fig:vqa_analysis}(c) demonstrates the impact of the number of demonstration examples on performance, revealing that the model achieves optimal performance with one demonstration example for the I$^2$L method. 

\noindent\textbf{Impact of Threshold for Selection.} I$^2$L performs well with complex images, while VT-ICL is effective for images that can be easily described by text. A score ranging from 1 to 4 with GPT-4V is used to rate the ease of describing an image. Figure~\ref{fig:vqa_analysis}(d) shows the threshold for choosing between I$^2$L or VT-ICL. For example, a threshold of 1.5 means that I$^2$L is used for scores below 1.5 and VT-ICL for scores above. Our findings suggest that 1.5 is the optimal threshold for the Mathvista dataset, and we use this as our default threshold in our main result table.


\section{Conclusion}
In this paper, we propose a novel approach called In-Image Learning (I$^2$L) to enhance the capabilities of GPT-4V. I$^2$L combines demonstration examples, visual cues, and instructions into a single image, providing an in-context learning experience.
I$^2$L excels at handling complex images, while interleaved visual-text in-context learning is better suited for images that can be easily described by text. To leverage the strengths of both methods for multimodal tasks, we propose I$^2$L-Hybrid which uses GPT-4V as a selector to determine the appropriate method for each multimodal data example in a given task. Through comprehensive experiments on MathVista, we demonstrate the effectiveness of our proposed method in complex reasoning tasks. We also examine the impact of factors such as image resolution and the positioning of demonstration examples, further highlighting the potential of I$^2$L.



\bibliography{colm2024_conference}
\bibliographystyle{colm2024_conference}

\clearpage
\appendix

\section{Appendix}

\subsection{Limitations}
This paper presents In-Image Learning as a means to enhance the capabilities of GPT-4V for multimodal tasks. One limitation of the proposed method is its sensitivity to the position of demonstrations. Additionally, we did not conduct experiments on open-source large multimodal models. In future work, we will explore methods to reduce the sensitivity to position in image demonstrations and investigate the implementation of in-image learning on more open-source large multimodal models.

\subsection{Demonstration Construction Example of MathVista}
\label{sec:appedix_demo}
\begin{figure*}[t]
    \centering
    \includegraphics[width=0.86\linewidth]{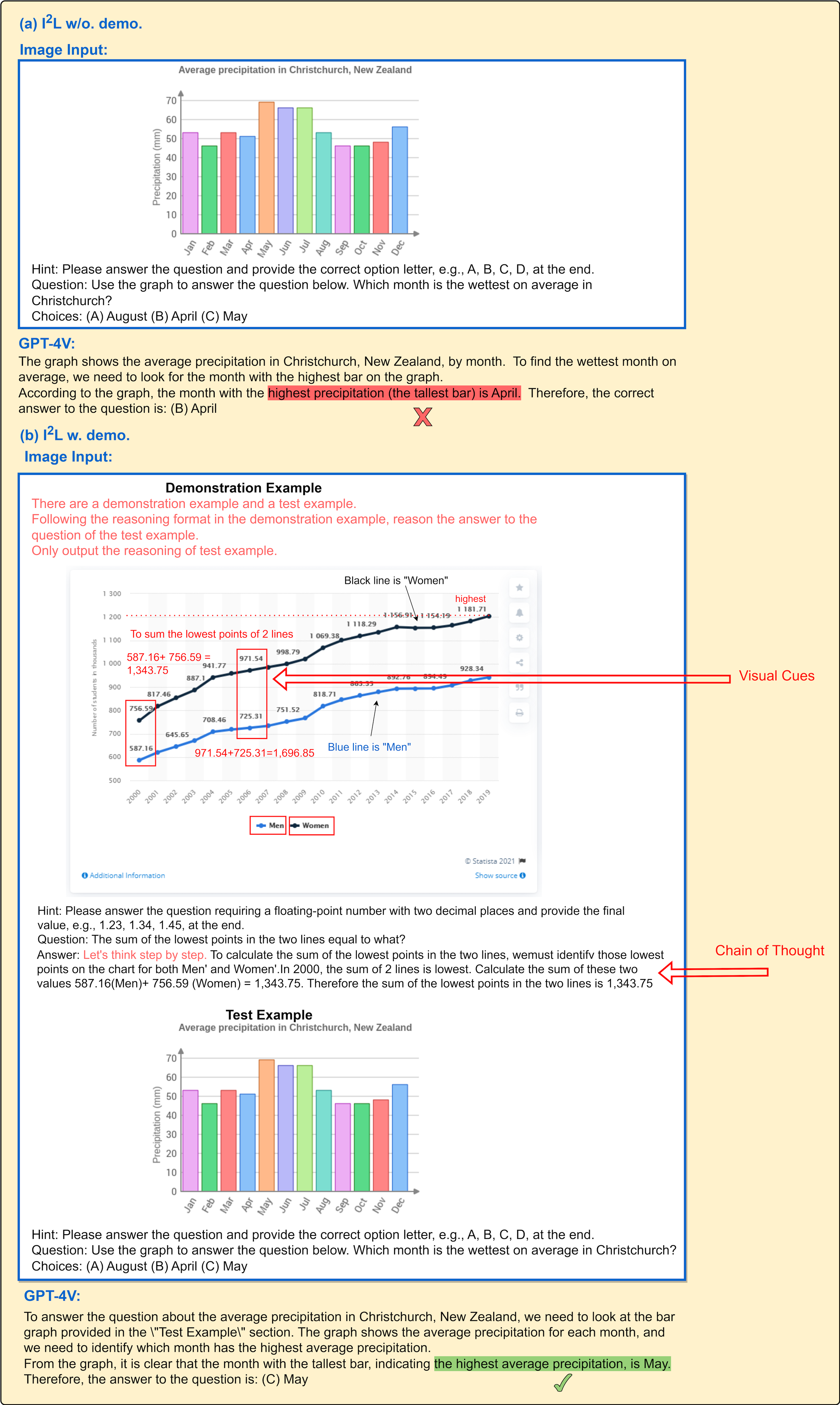}
    \caption{ Example of I$^2$L for MathVista. GPT4-V is not able to generate the correct answer without demonstrate. With adding visual cues and Chain-of-Thought rationales on the demonstration example, GTP4-V is able learn from the demonstration to solve the test example, even there is no additional information for the test example.}
    
    \label{fig:I2L_mathvista}
    \vspace{-5pt}
\end{figure*}

Figure \ref{fig:I2L_mathvista} illustrates an exemplar construction on MathVista dataset. Adhering to the prescribed construction criteria, we incorporate visual cues, such as boxes, to highlight pivotal components within the exemplar. It is imperative during the integration of visual cues and demonstrations to ensure that no essential objects in the original image are obscured, thus mitigating the risk of information loss. Additionally, a Chain-of-Thought rationale is included to bolster the model's capacity of reasoning. By leveraging this demonstration example, the model can engage in-context learning through image-only input, thereby facilitating accurate responses to test questions.

\begin{figure*}[t]
    \centering
    \includegraphics[width=1\linewidth]{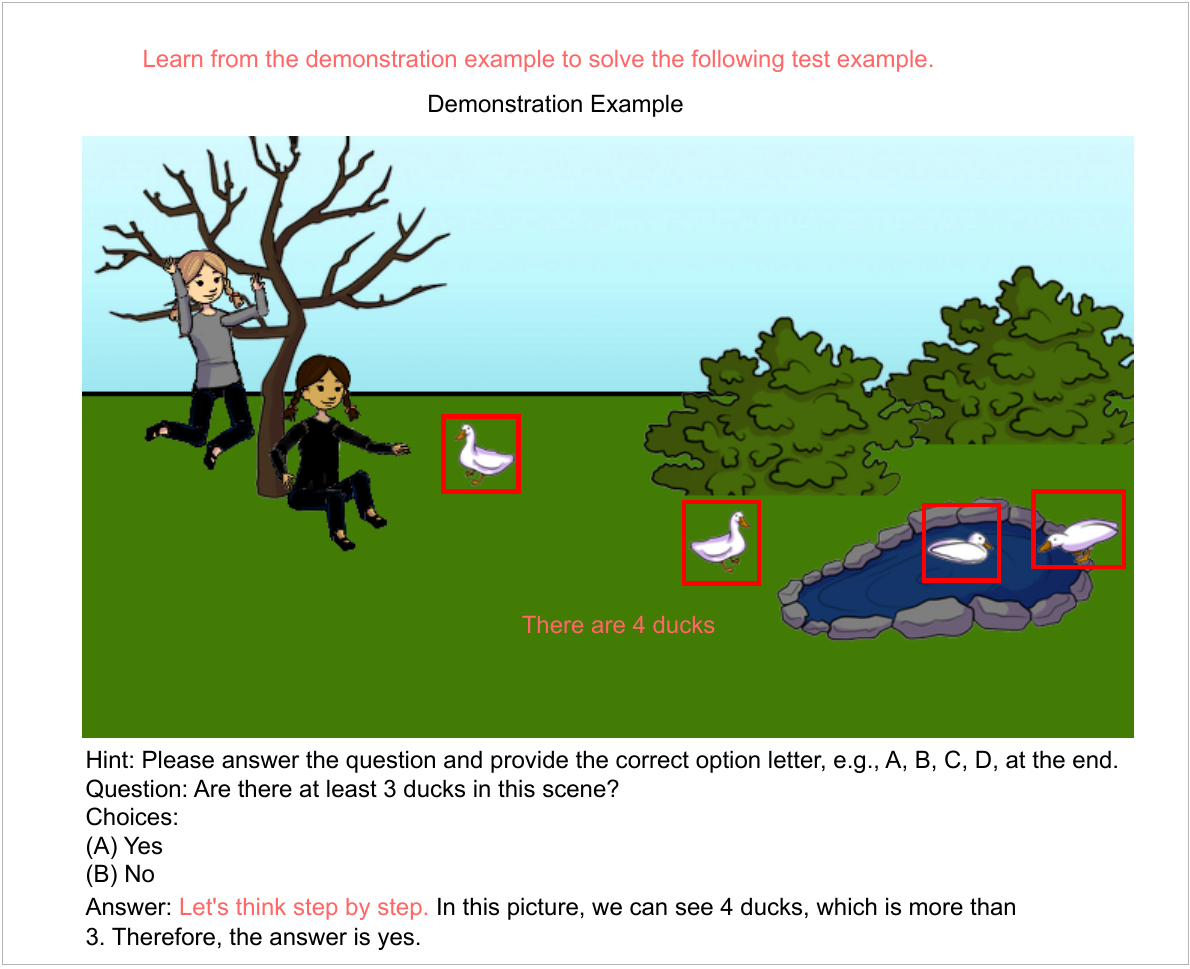}
    \caption{ Demonstration of abstract\_scene in mathvista.}
    
    \label{fig:demo_abstract_scene}
    \vspace{-5pt}
\end{figure*}

\begin{figure*}[t]
    \centering
    \includegraphics[width=1\linewidth]{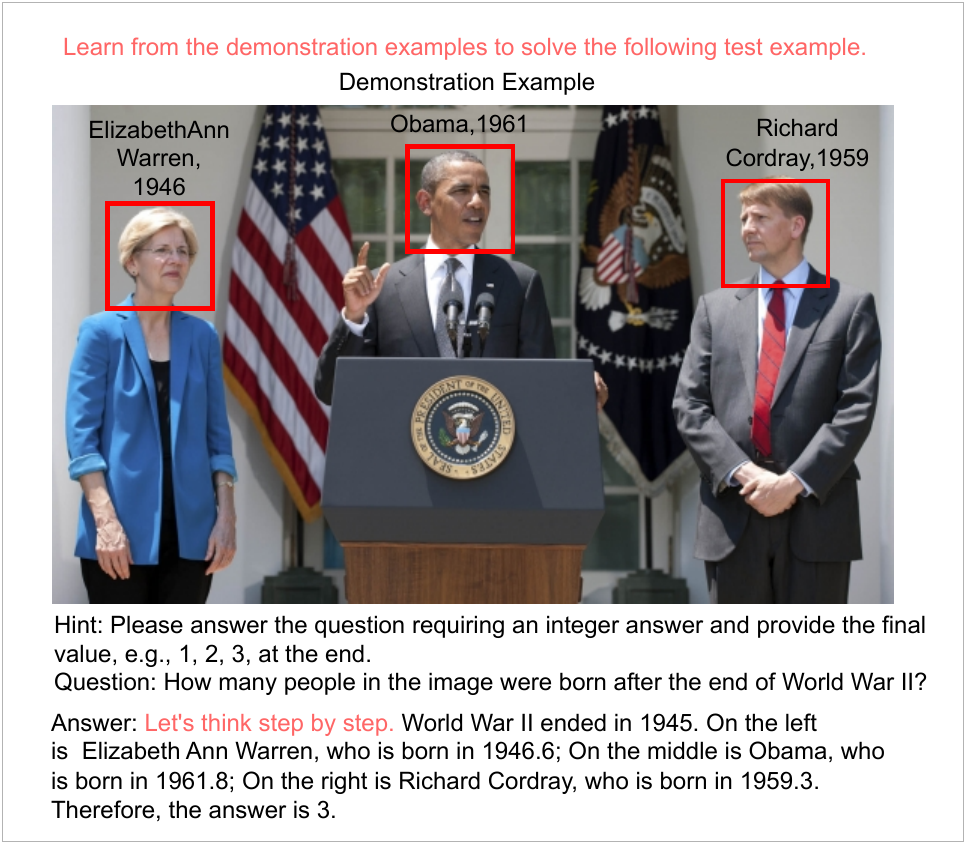}
    \caption{ Demonstration of natural\_image in mathvista.}
    
    \label{fig:demo_paper_case-natural_image}
    \vspace{-5pt}
\end{figure*}

\begin{figure*}[t]
    \centering
    \includegraphics[width=1\linewidth]{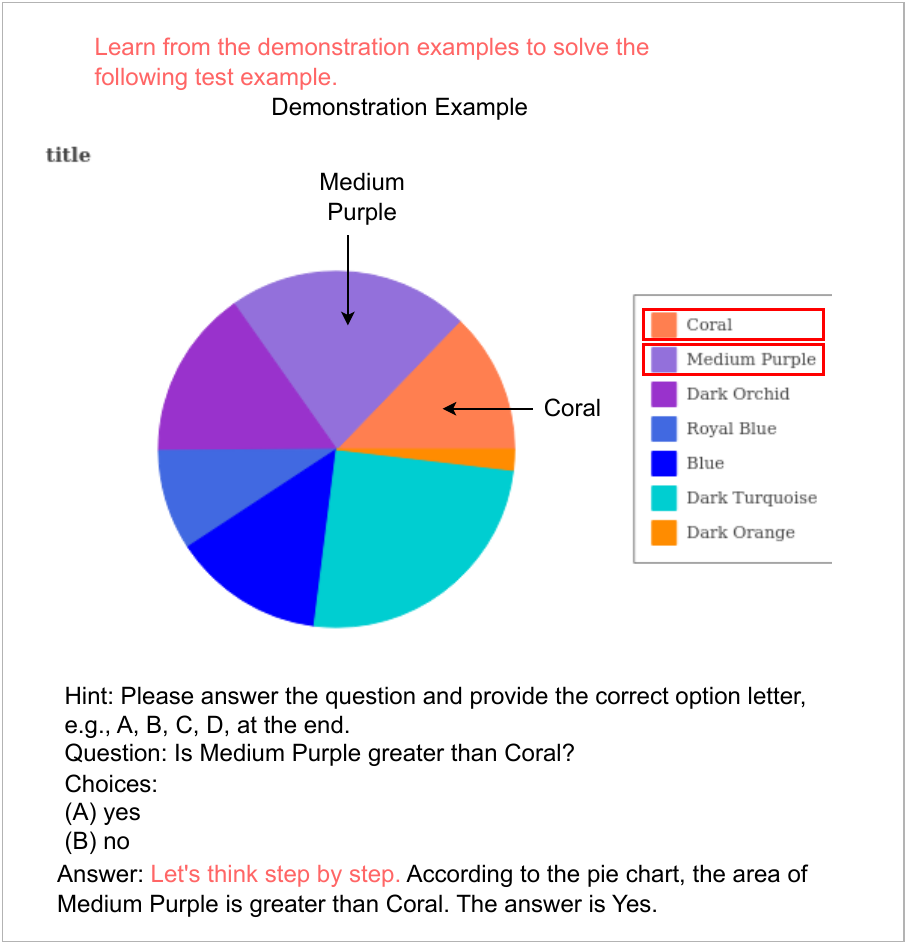}
    \caption{ Demonstration of pie\_chart in mathvista.}
    
    \label{fig:demo_paper_case-pie_chart}
    \vspace{-5pt}
\end{figure*}

\begin{figure*}[t]
    \centering
    \includegraphics[width=1\linewidth]{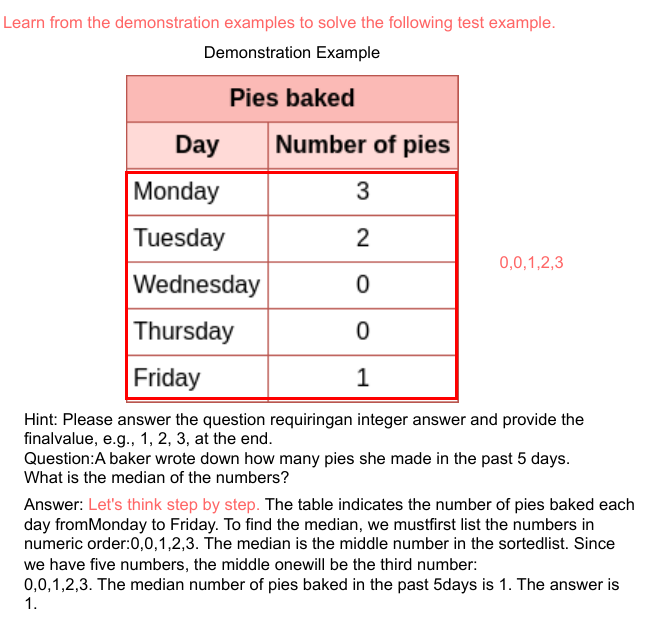}
    \caption{ Demonstration of table in mathvista.}
    
    \label{fig:demo_paper_case-table}
    \vspace{-5pt}
\end{figure*}

\begin{figure*}[t]
    \centering
    \includegraphics[width=1\linewidth]{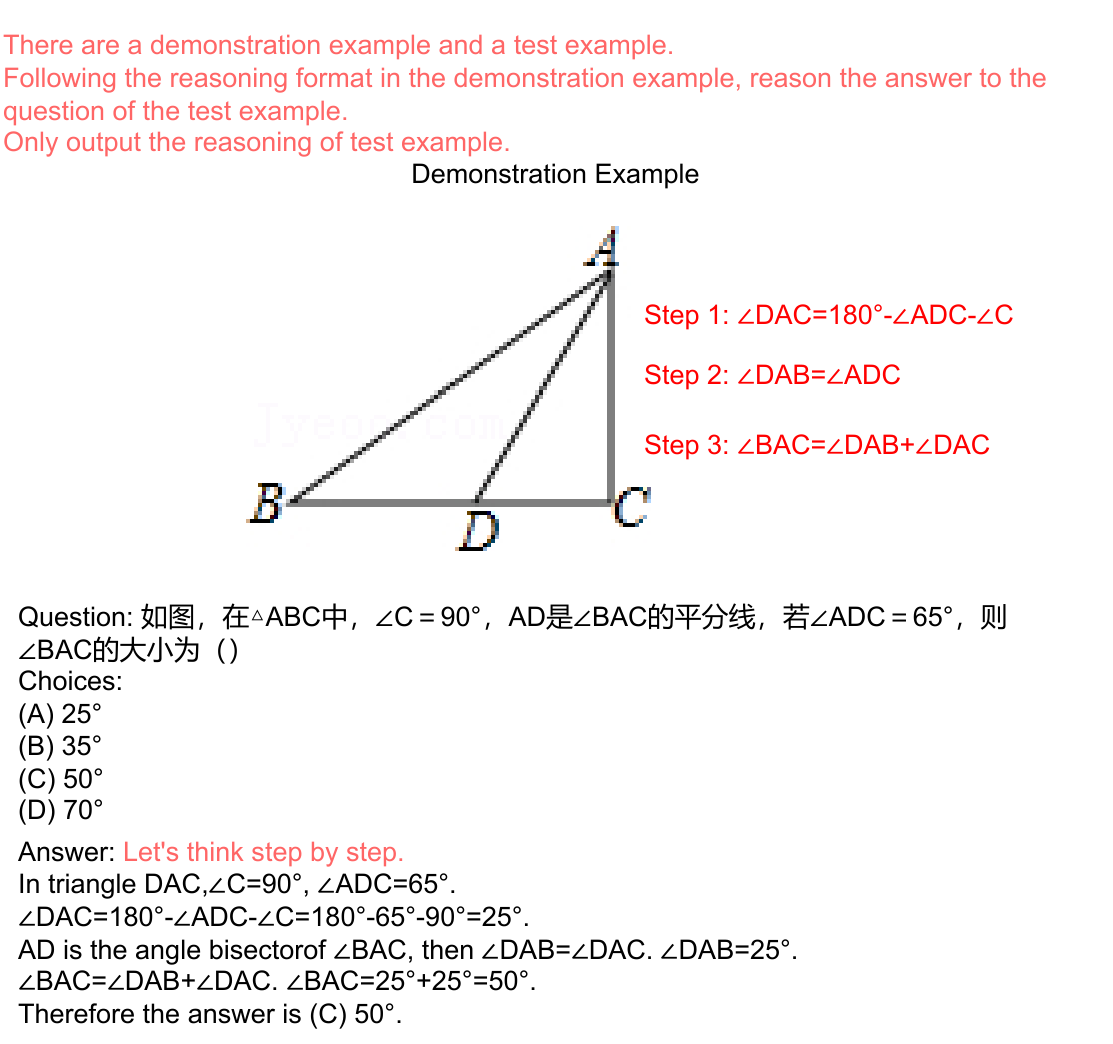}
    \caption{ Demonstration of geometry\_diagram in mathvista.}
    
    \label{fig:demo_paper_case-geometry_diagram}
    \vspace{-5pt}
\end{figure*}

To comprehensively show how we construct the demonstration examples, Figure \ref{fig:demo_abstract_scene}, Figure \ref{fig:demo_paper_case-geometry_diagram}, Figure \ref{fig:demo_paper_case-natural_image}, Figure \ref{fig:demo_paper_case-pie_chart}, and Figure \ref{fig:demo_paper_case-table} show the demonstration examples of different subsets of MathVista dataset.

\begin{figure*}[t]
    \centering
    \includegraphics[width=0.86\linewidth]{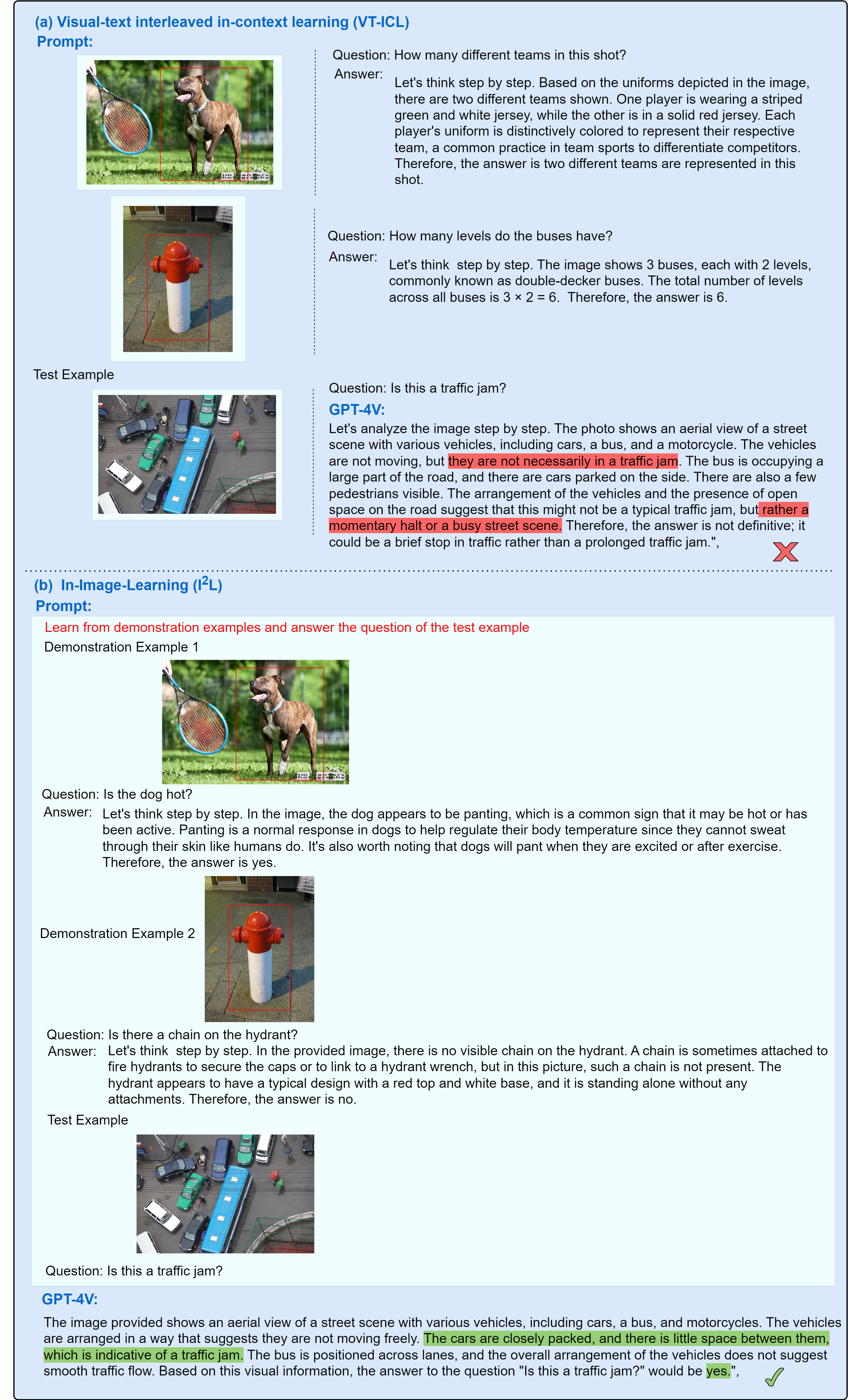}
    \caption{ An case of the yesorno task. (a): Input with image demonstrations and in-context-learning from demonstrations to solve the test question. (b): Input with image demonstrations and learning from demonstrations to solve the test question.}
    
    \label{fig:case_yesorno_I2L_VT-ICL}
    \vspace{-5pt}
\end{figure*}

\begin{figure*}[t]
    \centering
    \includegraphics[width=0.86\linewidth]{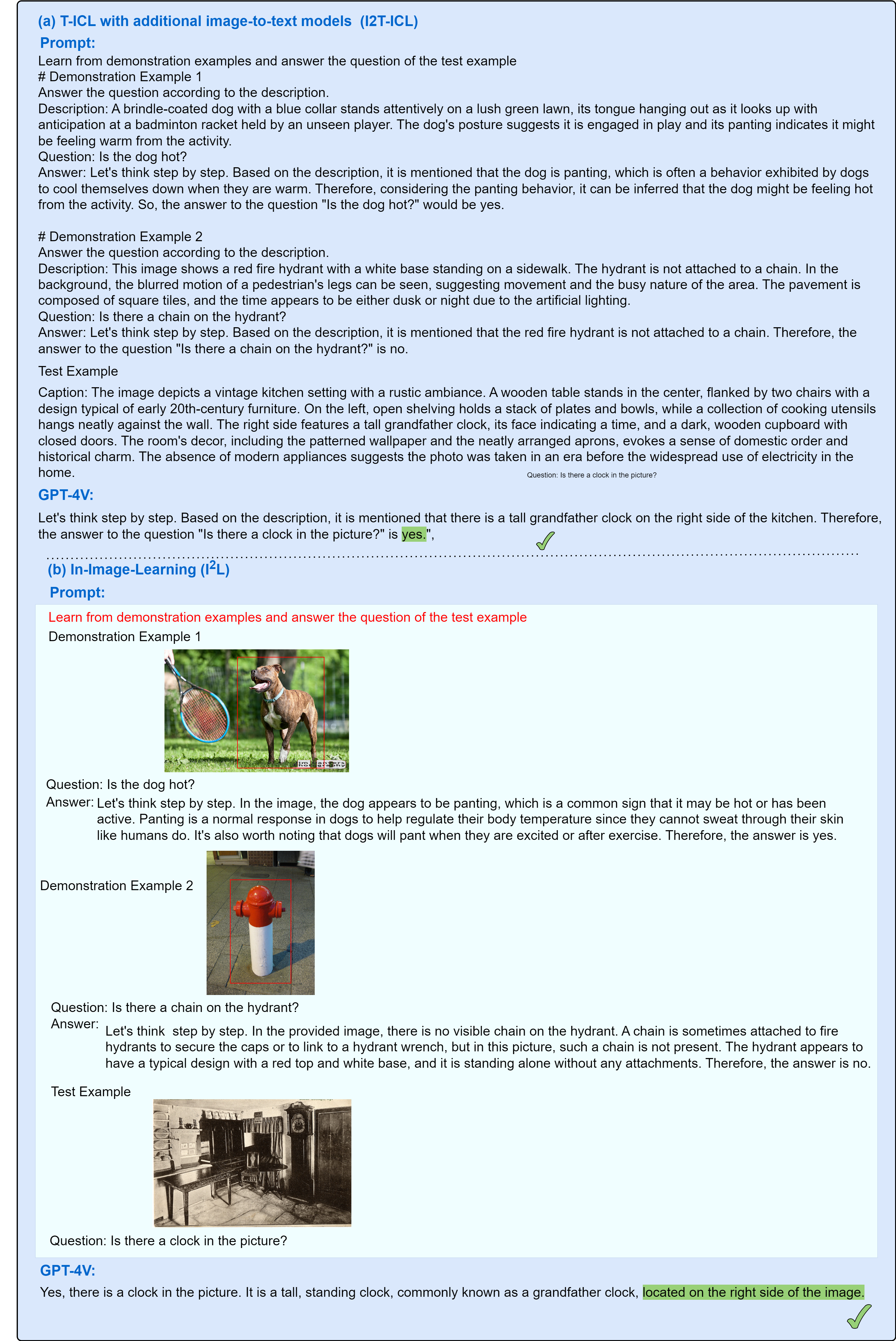}
    \caption{ An case of the yesorno task.
    (a): Input with Text-only in-context learning with additional image-to-text models to solve the test question. (b): Input with image demonstrations and in-context-learning from demonstrations to solve the test question.}
    
    \label{fig:case_yesorno_T-ICL_I2L}
    \vspace{-5pt}
\end{figure*}

\begin{figure*}[t]
    \centering
    \includegraphics[width=0.86\linewidth]{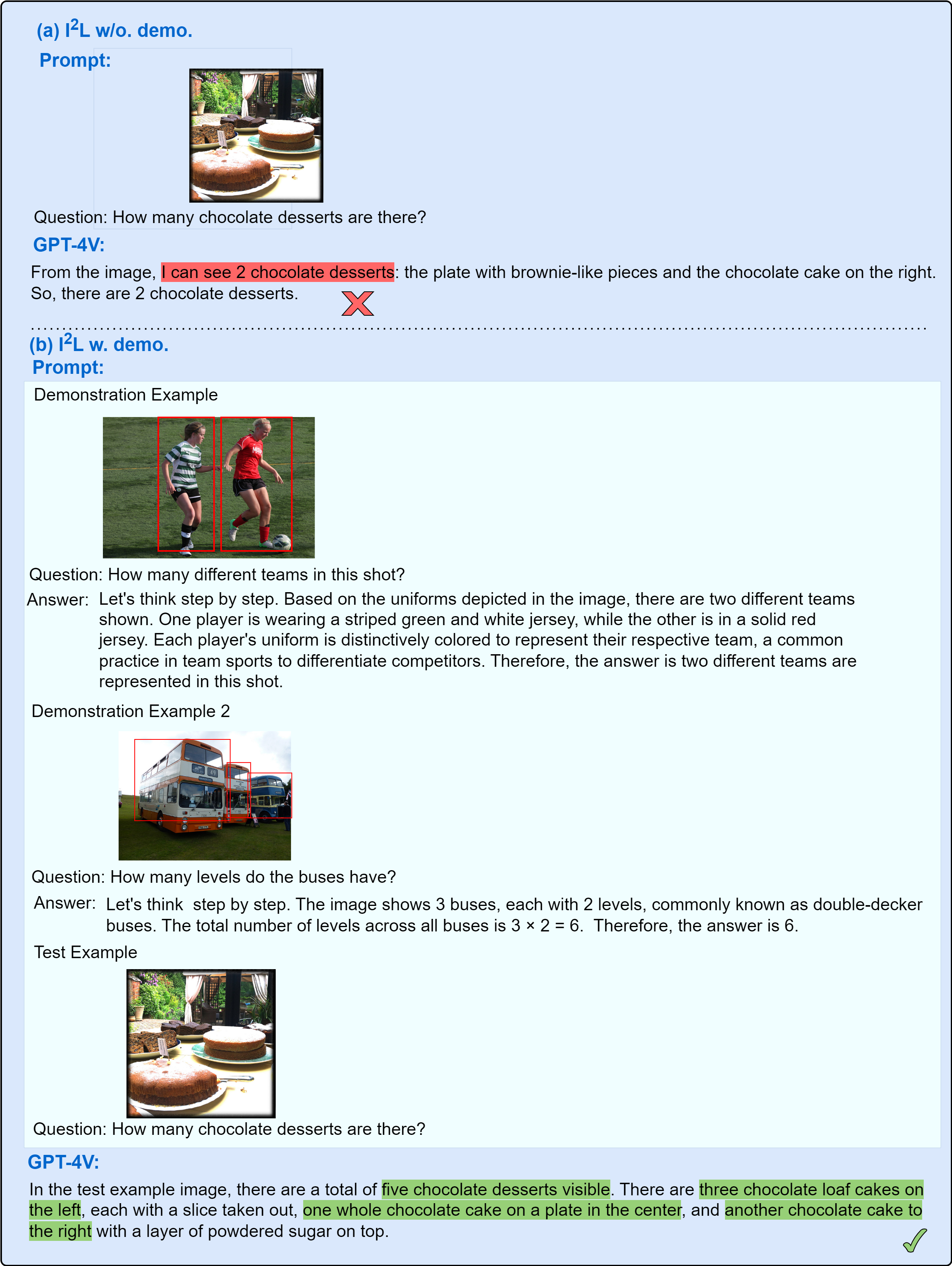}
    \caption{  An case of the yesorno task.
    (a) Input without demonstration. (b) Input with In-image learning from demonstrations to solve the test question.}
    
    \label{fig:case_yesorno_w_o_demo}
    \vspace{-5pt}
\end{figure*}

\begin{figure*}[t]
    \centering
    \includegraphics[width=1\linewidth]{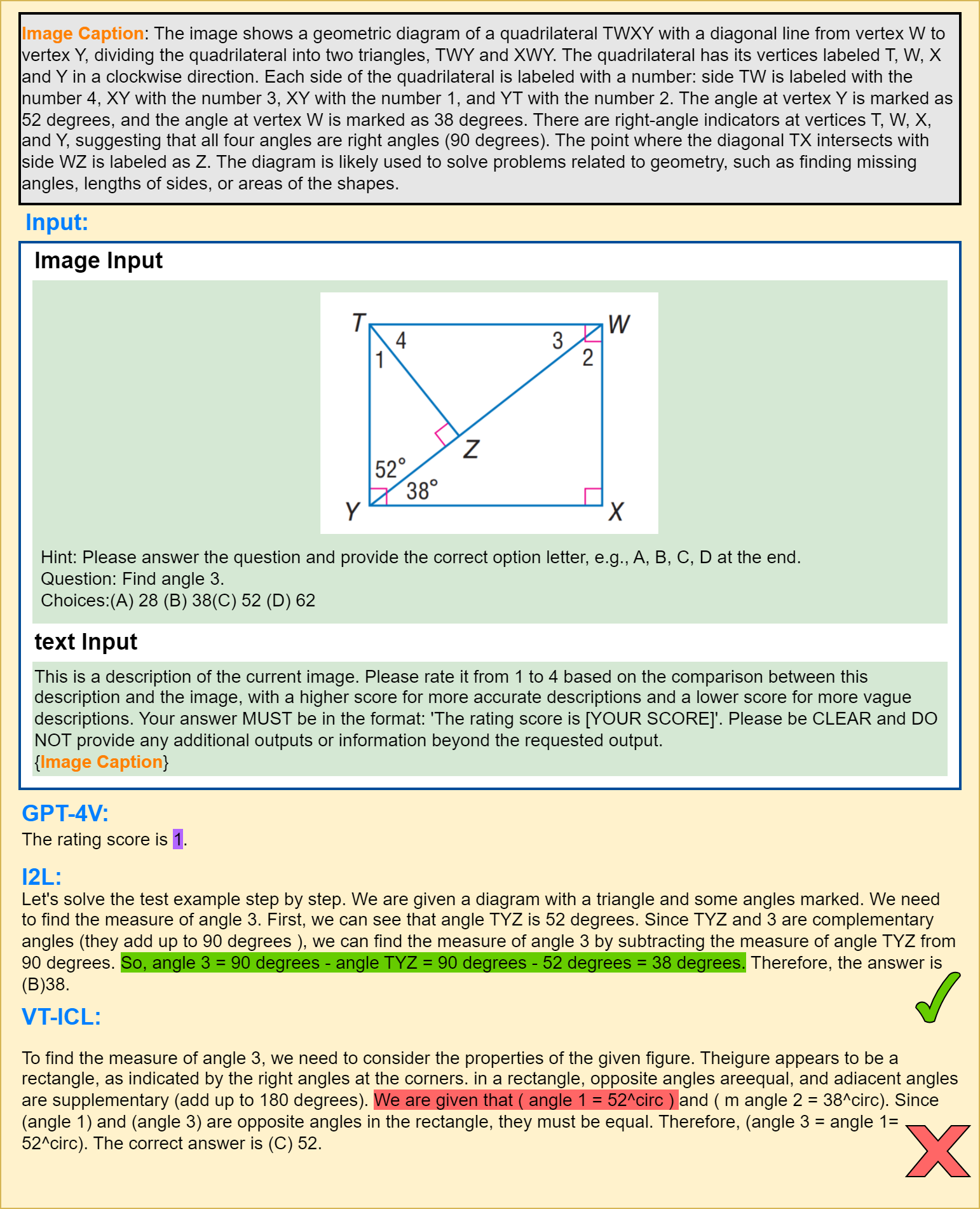}
    \caption{The selection methods of I$^2$L-Hybrid. GPT4-V gives the rate score 1 to the sample which means I$^2$L is used to solve the problem. And I$^2$L is able to correctly answer the question.}
    
    \label{fig:selector_case}
    \vspace{-5pt}
\end{figure*}

\subsection{Experiment Results of VQA}
\label{sec:appedix_vqa}
VQA contains open-ended questions about images for evaluating the LMM's ability to understand vision, language, and common sense. Instead of using the entire VQA dataset, we construct three subsets covering three types of questions, i.e., VQA(YorN), VQA(Counting), and VQA(others).  Each subset consists of randomly selected 50 examples.

In order to assess the general vision understanding capabilities of the proposed I$^2$L method, we conducted experiments on the Visual Question Answering (VQA) dataset. Table \ref{tab:vqa_result} provides an overview of the results obtained by I$^2$L, T-ICL-Img, and VT-ICL on the VQA dataset. Notably, the I$^2$L method surpasses T-ICL-Img and VT-ICL, achieving an average accuracy of 72\%. This outcome underscores the superior ability of I$^2$L to enhance the overall vision understanding capabilities of the models.


\begin{table}[h]
    \centering
    \resizebox{0.7\textwidth}{!}{
        \begin{tabular}{lccccc}
            \toprule
              & \textbf{VQA(YorN)} & \textbf{VQA(Counting)} & \textbf{VQA(others)} & \textbf{Average}\\
            \midrule
            \textbf{T-ICL-Img}   & 0.64 & 0.46 & 0.44 & 0.51\\
            \textbf{VT-ICL}  & \textbf{0.78} & 0.56 & 0.68 & 0.64 \\
            \textbf{I$^2$L}   & 0.76 & \textbf{0.66} & \textbf{0.74} & \textbf{0.72} \\
            \bottomrule
        \end{tabular}
    }
    \caption{In-image learning: Accuracy results for VQA(YorN), VQA(Counting) and VQA(others) datasets.}
    \label{tab:vqa_result}
\end{table}

\subsection{Experiment Results of HallusionBench}
\label{sec:appedix_hallusion}
We adhere to the methodology outlined in HallusionBench \cite{liu2023hallusionbench} for the evaluation of LLMs across various assessment metrics, including the Yes/No bias test, Consistency test, and Language and Vision Diagnosis. The results of these evaluations are summarized in Table \ref{tab:leaderboard-test}. Notably, both human evaluation and GPT-4-Assisted evaluation metrics are reported, in accordance with the protocol established by \citet{liu2023hallusionbench}. Upon examination of the table, it becomes apparent that the GPT-4 model continues to exhibit bias in automatic evaluations, as indicated by discrepancies between the GPT-4-Assisted evaluation and human evaluation results. Consequently, our analysis primarily focuses on the human evaluation outcomes, with GPT-4-Assisted results serving as supplementary evidence.

In the context of the Yes/No bias test, T-ICL achieves the most favorable Yes Percentage Difference (Pct. Diff) score of -0.02, while I$^2$L demonstrates the lowest False Positive Ratio (FP Ratio) of 0.23, as evidenced by both human evaluation and GPT-4-Assisted evaluation. In the Consistency test, T-ICL attains the highest Correct and Inconsistent score at 24.63\% and 53.62\%, respectively, while I$^2$L achieves the superior Wrong score at 13.04\%. Of particular note is the performance of the proposed I$^2$L model in the Language and Vision Diagnosis, wherein it surpasses other models with a Mixed score of 57.81\%. This outcome signifies that I$^2$L exhibits efficacy in mitigating language hallucination and visual illusion, thereby enhancing diagnostic accuracy in multimodal contexts.

\begin{table*}[t]
\vspace{-1em}
  \begin{center}
  \fontsize{12pt}{12pt}
  \resizebox{\textwidth}{!}{
  \begin{tabular}{lccccccccc}
    \toprule
    \multicolumn{2}{c}{} &  \multicolumn{2}{c}{\textbf{Yes/No Bias}} &  \multicolumn{3}{c}{\textbf{Consistency}} &  \multicolumn{3}{c}{{ \textbf{Language and Vision Diagnosis}
    }}  \\ \cmidrule(lr){3-4} \cmidrule(lr){5-7}  \cmidrule(lr){8-10}              
    \textbf{Method} & \textbf{Evaluation} & Pct. Diff$(\sim 0)$ & FP Ratio$(\sim 0.5)$ & Correct $\uparrow$
    & Inconsistent $\downarrow$ & Wrong $\uparrow$ & Lang. Hallus. & Visual Illus. & Mixed\\
    \midrule
    Random Chance & GPT-4-Assisted & 0.08 & \bf0.57 & 18.20 & 57.51 & 24.28 & - & - & - \\
     \midrule
    \multirow{2}{*}{LLaVA-1.5~\cite{liu2023improved}}  & Human  & 0.27 &   0.76   &          25.43      &     42.49   &         32.08          & 25.63 & 51.42 & 22.95  \\
    & GPT-4-Assisted & 0.26   &   0.75   &          \textbf{24.86}        &             \textbf{45.38}          &         29.77          & 26.71 & 51.09 & 22.20 \\
    \midrule 
    BLIP2-T5~\cite{li2023blip}  & GPT-4-Assisted & \bf0.08 & 0.58 & {20.52} & 59.54 & \textbf{19.94} & 41.64 & 40.44 & 17.92\\
     \midrule
    Qwen-VL~\cite{bai2023qwen} & GPT-4-Assisted & 0.12 & 0.60 & 6.65 & \textbf{50.29} & 43.06 & 0.87 & \textbf{88.06} & 11.06\\
    Open-Flamingo~\cite{awadalla2023openflamingo}  & GPT-4-Assisted & 0.33 & 0.77 & 11.27 & 59.83 & 28.90 & 30.07 & 48.06 & 21.87\\ 
    MiniGPT5~\cite{zheng2023minigpt}  & GPT-4-Assisted & 0.28   &   0.71   &           9.83           &             56.36             &          33.82          & 10.09 & 73.44 & 16.47 \\
    MiniGPT-4~\cite{zhu2023minigpt} & GPT-4-Assisted & 0.19 & 0.65 & 10.12 & 57.80 & 32.08 & 23.59 & 56.55 & 19.86\\
    InstructBLIP~\cite{instructblip} & GPT-4-Assisted &-0.13 & 0.38 & 10.12 & 68.50 & \textbf{21.39} & 29.29 & 54.53 & 16.18\\
    BLIP2~\cite{li2023blip} & GPT-4-Assisted & 0.18 & 0.65 & 12.43 & 63.01 & 24.57 & 39.14 & 43.45 & 17.41\\
    mPLUG\_Owl-v2~\cite{ye22023mplug} & GPT-4-Assisted & 0.25 & 0.77 & 19.94 & 58.09 & 21.97 & 28.24 & 50.42 & 21.34\\
    mPLUG\_Owl-v1~\cite{ye2023mplug} & GPT-4-Assisted & 0.32 & 0.79 & 10.40 & 60.12 & 29.48 & 3.95 & \textbf{78.36} & 17.69\\
    LRV\_Instruction~\cite{liu2023aligning}  & GPT-4-Assisted & 0.26 & 0.73 & 13.01 & 53.47 & 33.53 & 4.49 & \textbf{76.47} & 19.04\\
     \midrule
    GIT~\cite{wang2022git} & GPT-4-Assisted & \textbf{0.04} & \textbf{0.53} & 6.36 & 53.76 & 39.88 & 30.90 & 58.30 & 10.80 \\
     \midrule
     \multirow{2}{*}{GPT-4V~\cite{openai2023gpt4vision}} & Human  & 0.07 & 0.60 & 44.22 & 32.66 & 23.12 & 21.86 & 46.17 & 31.97   \\
    & GPT-4-Assisted  & {0.06} & {0.58} & \textbf{39.88} & \textbf{38.15} & {21.97} & 22.19 & 45.66 & 32.14   \\
    \midrule
    \multirow{2}{*}{1-shot T-ICL (GPT-4V)} & Human  & {-0.02} & 0.46 & 24.63 & {53.62} & 26.73 & 49.01 & 58.75 & 56.39   \\
    & GPT-4-Assisted  & ~\textbf{-0.03} & \textbf{0.45} & \textbf{24.63} & 57.97 & 17.39  & \textbf{54.90} & 60.00 & \textbf{58.76}   \\
    \multirow{2}{*}{1-shot VT-ICL (GPT-4V)} & Human  & -0.10 & 0.37 & {20.28} & 66.66 & 13.04 & 49.01 & 59.37 & 56.87   \\
    & GPT-4-Assisted  & -0.10 & 0.37 & {20.28} & 66.66 & 13.04 & \textbf{49.01} & 58.12 & \textbf{55.92}   \\
    \multirow{2}{*}{1-shot I$^2$L (GPT-4V)} & Human  & -0.14 &  0.31 & 24.63 & 63.76 & {11.59} & 62.74&  {60.62} & {61.13}\\
    & GPT-4-Assisted  & -0.14 &  0.32 &  23.18 &  65.21&   \textbf{11.59} &  \textbf{62.74} &  59.37 &  \textbf{60.18}  \\ 
 
  \bottomrule
\end{tabular}}
\end{center}
\vspace{-3mm}
\caption{\textbf{Analytical Evaluation Results on HallusionBench with various LVLMs:} \textit{Pct. Diff} ranges from [-1, 1]. The model is more biased when \textit{Pct. Diff} is close to -1 or 1. \textit{FP Ratio} ranges from [0, 1]. The model is more robust when \textit{FP Ratio} is close to 0.5. All the other metrics are presented in \%, and the full score is 100\%. We highlight the Top 3 models with the GPT-4-assisted evaluation. }
\label{tab:leaderboard-test}
\vspace{-3mm}
\end{table*}

\subsection{Case Study}
To qualitatively show the performance of the proposed method I$^2$L and baseline methods, we randomly choose a few examples from VQA and MathVista datasets. Figure \ref{fig:case_yesorno_I2L_VT-ICL}, Figure \ref{fig:case_yesorno_T-ICL_I2L}, and Figure \ref{fig:case_yesorno_w_o_demo} show the generated responses of different methods.

Finally, Figure \ref{fig:selector_case} shows an example of our I$^2$L-Hybrid selection method.

\end{document}